\documentclass[letterpaper, 10 pt, conference]{ieeeconf}  

\IEEEoverridecommandlockouts                              

\usepackage{marvosym}  

\title{
\large \bf
Transformer-Based Token Fusion and Dynamic Graph Planning for Audio-Visual Navigation$\dagger$
}

\author{
Shaohang Wu$^{1,2,3}$, and Yinfeng Yu$^{1,2,3}$$^{,\mbox{\Letter}}$%
\thanks{\small $\dagger$This research was financially supported by the National Natural Science Foundation of China (Grant No. 62463029).}
\thanks{\small $^1$School of Computer Science and Technology, Xinjiang University, Urumqi 830017, China.}
\thanks{\small $^2$Joint Research Laboratory for Embodied Intelligence, Xinjiang University.}
\thanks{\small $^3$Joint International Research Laboratory of Silk Road Multilingual Cognitive Computing, Xinjiang University.}
\thanks{\small $^{\mbox{\Letter}}$Yinfeng Yu is the corresponding author (Email: yuyinfeng@xju.edu.cn).}%
}

\usepackage{cite}
\usepackage{amsmath,amssymb,amsfonts}
\usepackage{algorithmic}
\usepackage{graphicx}
\usepackage{textcomp}
\usepackage{xcolor}
\usepackage{multirow}
\usepackage{booktabs}

\begin{document}

\maketitle
\thispagestyle{empty}
\pagestyle{empty}

\begin{abstract}
Audio-Visual Navigation (AVN) requires an agent to localize and navigate toward a continuously vocalizing target relying solely on visual observations and acoustic cues. Currently, systems lack the ability to adaptively correct and replan when faced with incomplete or misleading visual perception. Furthermore, relying on physical collisions to compensate for missing visual information results in inefficient and unsafe navigation, whereas existing methods are overly dependent on passive visual perception. To address these issues, we propose the Transformer-based Token Fusion and Dynamic Graph Planning (TDGP) model, which incorporates high-level perception layers and leverages the Transformer model to fuse multimodal cues for precise local planning. Next, a low-level planning layer was designed that uses physical collision penalties to remove edges that collide with the map in real time and apply corresponding penalties, forcing the agent to automatically re-plan to compensate for the lack of visual information. Experiments show that our TDGP model outperforms baseline models on the Replica and Matterport3D(MP3D) datasets, and that the model’s sound enhancement strategy significantly improves generalization in unheard acoustic scenarios.


\end{abstract}

\section{INTRODUCTION}

Human navigation in the physical world often relies on the coordination of vision and hearing to locate targets. With the development of embodied intelligence, audio-visual Navigation \cite{b2,b3} has gradually become a hot topic of research. Compared to visual perception alone, audio-visual navigation relies on both visual and auditory cues to locate sound sources, significantly enhancing navigation capabilities in complex environments. This integration of visual and auditory perception enhances the practical value of intelligent agents in dynamic \cite{yu2021weavenet,wang2025modality}, complex scenarios such as home services and emergency search-and-rescue operations.

\begin{figure}[t]
\includegraphics[width=1\columnwidth]{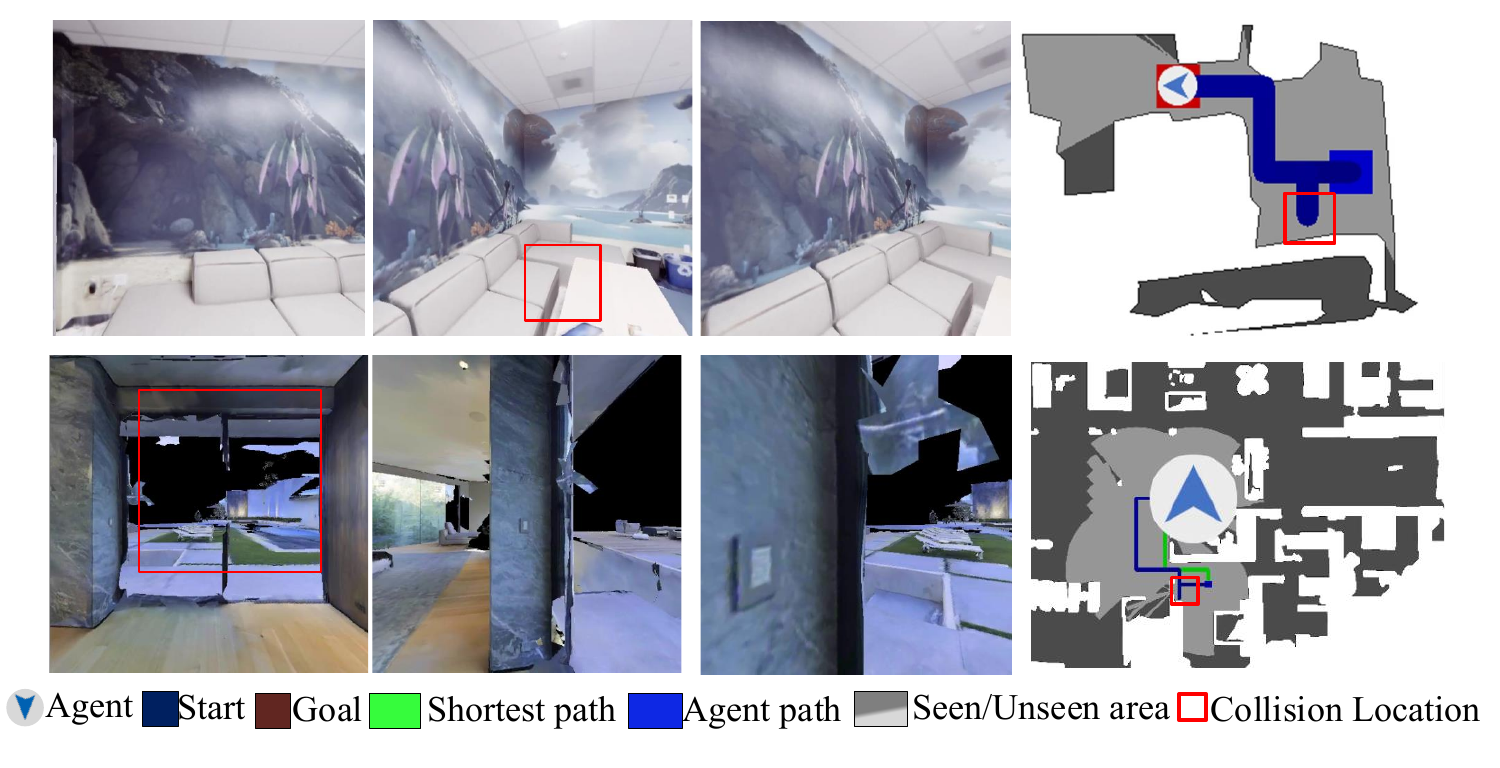} 
\caption{Passive visual perception is inherently limited and potentially misleading. For example, narrow hallways and glass doors.}
\label{fig:fig6}
\end{figure}

In complex real-world environments, relying solely on vision-based occupancy maps has two limitations, as shown in Fig. \ref{fig:fig6}: An agent's perception is limited by a narrow, self-centered field of view and lacks a global context; at the same time, transparent obstacles or abnormal lighting often obscure geometric details, thereby generating false passability signals. Traditional models often place blind trust in this type of flawed data, leading to planning failures. We believe that, in the absence of a preconstructed environmental map, physical collisions provide a reliable source of geometric information. However, frequent collisions are neither safe nor efficient; relying solely on passive post-collision recovery is insufficient for building a stable navigation system.

In recent years, Chen et al. \cite{b2} first introduced audio-visual navigation as an end-to-end multimodal deep reinforcement learning framework that encodes visual and auditory inputs with CNNs, fuses temporal cues via GRUs, and outputs low-level navigation actions. Subsequent studies proposed AV-WaN \cite{b5} for hierarchical navigation with dynamic waypoints and acoustic memory, semantic audio-visual navigation \cite{b23} for indirect sound sources, dynamic audio-visual navigation \cite{b17}, and dynamic multi-target fusion for efficient audio-visual navigation \cite{yu2025dynamic}. However, these methods generally map raw inputs directly to actions \cite{b2,b4,b5,b6,b26}, making them less effective under perceptual uncertainty.

\begin{figure*}[htb!] 
\centering
\includegraphics[width=\textwidth]{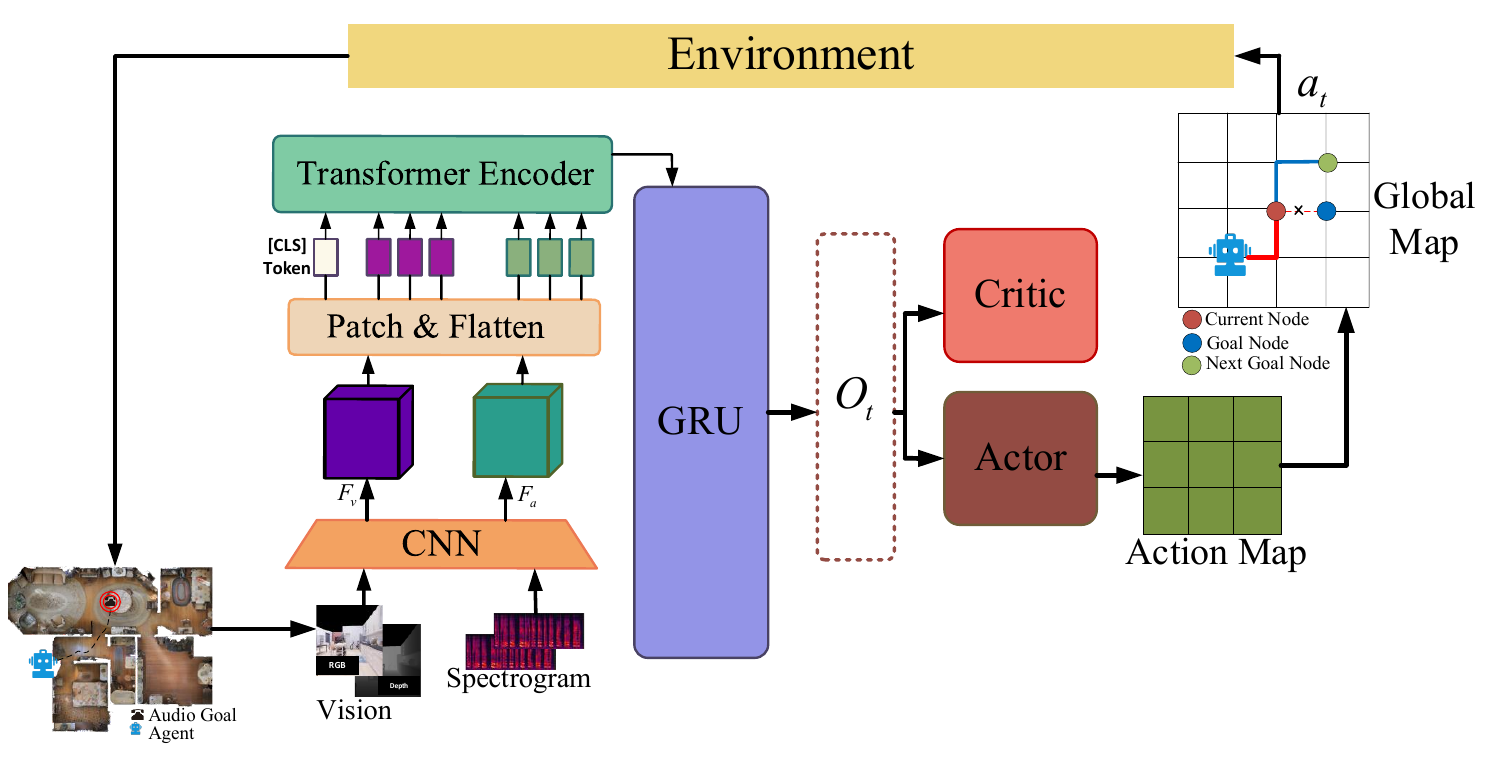} 
\caption{Transformer-based Token Fusion and Dynamic Graph Planning (TDGP) model. Through the Tokenization Fusion Mechanism in High-Level Transformers (TFM), the fused states are processed by a $\text{GRU}$, ultimately determining a local navigation target on a $3 \times 3$ action grid. This target is then passed to the Low-level Collision-Penalty Path Planner (CPP).}
\label{fig:fig2}
\end{figure*}

To address these issues, we propose the Transformer-based Token Fusion and Dynamic Graph Planning (TDGP) model, which decouples high-level decision-making from low-level control. To address the issue of insufficient integration between visual and auditory information, we have designed a tokenization fusion mechanism (TFM) for high-level transformers, which leverages TFM to achieve deep audio-visual alignment. However, due to the inherent uncertainty in visual perception, navigation can only be achieved through collision-guided auditory cues. To address this issue, we propose a low-level collision-penalty path planner(CPP) and design a collision penalty mechanism to mitigate the risks and inefficiencies associated with physical collisions for the agent. This design allows the high-level policy network to focus on decision-making regarding the target direction, while the low-level policy focuses on path selection. We have also incorporated audio data augmentation strategies into the SoundSpaces simulator to improve the model's robustness. Ultimately, TDGP was trained using PPO \cite{b12}, and the model demonstrated excellent generalization capabilities in unheard complex scenarios.

The main contributions of this paper are as follows:
\begin{itemize}
    \item We propose the Tokenization Fusion Mechanism in High-Level Transformers, which achieves deep alignment between audio and visual features. 
    \item We propose the Low-level Collision-Penalty Path Planner. When an agent collides, the system automatically removes the corresponding edge from the navigation map and applies a collision penalty.
    \item We propose the Transformer-based Token Fusion and Dynamic Graph Planning model, which decouples high-level perception strategies from low-level path planning. The model achieves state-of-the-art performance on the Replica and MP3D datasets, and in terms of sound enhancement strategies, it improves generalization in unheard acoustic scenarios.
\end{itemize}
The remainder of this paper is organized as follows: Section~II reviews related work, Section~III describes TDGP, Section~IV presents experiments, and Section~V concludes the paper.

\section{Related work}

\subsection{Hierarchical Reinforcement Learning and Planning for Audio-Visual Navigation}
To simplify long-term audio-visual navigation (AVN) \cite{b2}, we employ hierarchical reinforcement learning \cite{b5} with a high-level policy and a low-level planner. Unlike previous methods \cite{b5}, our high-level policy is invoked at every environmental time step, enabling the agent to update local goals based on the latest maps and sounds.

\subsection{Data Augmentation in Audio-Visual Navigation}
Improving model robustness in complex, noisy, and dynamic environments remains a core challenge for embodied agents \cite{YinfengICLR2022saavn,yang2026beyond,yu2025dope}. Previous studies have addressed this issue through moving object tracking \cite{b15} and multi-agent collaboration \cite{yu2023measuring,zhang2025advancing}. For data augmentation, Younes et al. \cite{b17} adopted a CNN-RNN-based framework. In contrast, we propose a Transformer-based framework that goes beyond recurrent architectures and provides greater flexibility for token-level integration of enhanced multimodal features.

\begin{table*}[htbp]
\centering
\caption{Experimental results trained on the heard and unheard splits of the Telephone dataset from the Replica and Matterport3D datasets.}
\label{tab:tab1}
\resizebox{\textwidth}{!}{
\begin{tabular}{@{}l ccc ccc ccc ccc@{}}
\toprule
\multirow{2}{*}{\textbf{Model}} & \multicolumn{6}{c}{\textbf{Replica}} & \multicolumn{6}{c}{\textbf{Matterport3D}} \\
\cmidrule(lr){2-7} \cmidrule(lr){8-13}
 & \multicolumn{3}{c}{\textit{Heard}} & \multicolumn{3}{c}{\textit{Unheard}} & \multicolumn{3}{c}{\textit{Heard}} & \multicolumn{3}{c}{\textit{Unheard}} \\
\cmidrule(lr){2-4} \cmidrule(lr){5-7} \cmidrule(lr){8-10} \cmidrule(lr){11-13}
 & SPL$\uparrow$ & SR$\uparrow$ & SNA$\uparrow$ & SPL$\uparrow$ & SR$\uparrow$ & SNA$\uparrow$ & SPL$\uparrow$ & SR$\uparrow$ & SNA$\uparrow$ & SPL$\uparrow$ & SR$\uparrow$ & SNA$\uparrow$ \\
\midrule
Random Agent~\cite{b5} & 4.9 & 18.5 & 1.8 & 4.9 & 18.5 & 1.8 & 2.1 & 9.1 & 0.8 & 2.1 & 9.1 & 0.8 \\
Direction Follower~\cite{b5}  & 54.7 & 72.0 & 41.1 & 11.1 & 17.2 & 8.4 & 32.3 & 41.2 & 23.8 & 13.9 & 18.0 & 10.7 \\
Frontier Waypoints~\cite{b5} & 44.0 & 63.9 & 35.2 & 6.5 & 14.8 & 5.1 & 30.6 & 42.8 & 22.2 & 10.9 & 16.4 & 8.1 \\
Supervised Waypoints~\cite{b5} & 59.1 & 88.1 & 48.5 & 14.1 & 43.1 & 10.1 & 21.0 & 36.2 & 16.2 & 4.1 & 8.8 & 2.9 \\
Gan et al.~\cite{b4} & 57.6 & 83.1 & 47.9 & 7.5 & 15.7 & 5.7 & 22.8 & 37.9 & 17.1 & 5.0 & 10.2 & 3.6 \\

AV-Nav~\cite{b2}          & 74.4 & 91.4 & 48.1 & 34.7 & 50.9 & 16.7 & 54.3 & 67.7 & 31.3 & 21.9 & 33.5 & 10.4 \\
AGSA~\cite{li2025audio} & 75.5 & 93.2 & 52.0 & 36.6 & 48.3 & 22.4 & 54.1 & 70.0 & 30.0 & 36.5 & 26.2 & 13.1 \\
\textbf{TDGP (Ours)}  & \textbf{89.2} & \textbf{98.7} & \textbf{73.7} & \textbf{41.7} & \textbf{60.0} & \textbf{26.2} & \textbf{62.6} & \textbf{73.7} & \textbf{54.2} & \textbf{36.1} & \textbf{42.2} & \textbf{25.9} \\
\bottomrule
\end{tabular}%
}
\end{table*}

\section{Method}
We propose the Transformer-based Token Fusion and Dynamic Graph Planning (TDGP) model, which aims to decouple navigation tasks and establish a two-layer policy. As shown in Fig. \ref{fig:fig2}, the system consists of two collaborative modules: the Tokenization Fusion Mechanism in High-Level Transformers (TFM) and the Low-level Collision-Penalty Path Planner (CPP).

\subsection{Tokenization Fusion Mechanism in High-Level Transformers} The Tokenization Fusion Mechanism in High-Level Transformers (TFM) is designed to fuse audio-visual sensor data streams to identify local navigation targets on a $3 \times 3$ grid. This functionality is implemented through a three-stage pipeline: multimodal feature extraction, transformer token fusion, and GRU-based temporal decision-making.

\paragraph{Multimodal Feature Extraction}
Multimodal features are extracted in parallel. Visual features $F_v$ are extracted from visual inputs using a dedicated visual encoder~\cite{fu2025fsdenet}, while acoustic features $F_a$ are extracted simultaneously from binaural audio spectrograms using an audio encoder.

\paragraph{Transformer Token Fusion}
We employ a Transformer architecture for multimodal fusion by tokenizing features from different modalities. For visual features $F_v$ and acoustic features $F_a$, we adopt a patch embedding strategy similar to that used in the Vision Transformer (ViT) \cite{b9}. We use a 2D convolutional layer as the patch embedding module, with both the kernel size and stride set to $P$. This operation linearly projects each $P \times P$ feature patch into a single token, with the embedding dimension uniformly set to $D_{\text{embed}}$ ($D_{\text{embed}}=128$). Consequently, the feature maps are converted into token sequences for the visual and auditory modalities, denoted as $\mathbf{T}_t^{\text{vis}}$ and $\mathbf{T}_t^{\text{aud}}$, respectively. Additionally, a learnable [CLS] token, $T_{\text{cls}}$, is introduced to aggregate global information across the sequence. The Transformer’s input sequence is constructed by concatenating all available tokens, which are then fed into the standard Transformer encoder module $\mathbf{E}_{\text{Trans}}$:
\begin{equation}
\mathbf{T}_t = [T_{\text{cls}} ; \mathbf{T}_t^{\text{vis}} ; \mathbf{T}_t^{\text{aud}}],
\end{equation}
\begin{equation}
\mathbf{c}_t  = \mathbf{E}_{\text{Trans}}(\mathbf{T}_t),
\end {equation}
Here, $\mathbf{c}_t \in \mathbb{R}^{D_{\text{embed}}}$ denotes the global context vector, which integrates information from all modalities. The Transformer’s multi-head attention mechanism \cite{b19,yu2025dgfnet,zhang2025iterative} facilitates deep information exchange between tokens of different modalities, thereby enabling effective cross-modal information fusion.

\paragraph{GRU-based Temporal Decision-Making}
The temporal decision module extracts the [CLS] token from the Transformer’s output to obtain the global context vector $\mathbf{c}_t$ for the current state, which integrates information from all modalities. Subsequently, this vector $\mathbf{c}_t$ is fed into a Gated Recurrent Unit (GRU) \cite{b11}, combined with the previous frame’s hidden state $h_{t - 1}$, to update the current hidden state $h_t$ and capture temporal dependencies. Finally, the module outputs a probability distribution $\pi(\mathbf{g}_t | h_t)$ over a $3 \times 3$ local grid (corresponding to 9 potential navigation targets $\mathbf{g}_t$), where $\pi(\mathbf{g}_t | h_t)$ represents the probability of selecting the local target $\mathbf{g}_t$ given the final hidden state $h_t$.

\subsection{Low-level Collision-Penalty Path Planner}
The Low-level Collision-Penalty Path Planner (CPP) transforms the local goal $\mathbf{g}_t$ from the high-level policy $\pi$ into a single atomic action $\mathbf{a}_t$. It maintains a dynamic navigation graph $\mathcal{G}_t=(V, E_t)$, where $V$ is the set of discrete navigable nodes and $E_t$ is the set of valid edges at time $t$.

\begin{table}[htbp]
\centering
\caption{Ablation study of TFM and CPP.}
\label{tab:tab3}
\renewcommand{\arraystretch}{1.2}

\resizebox{\columnwidth}{!}{%
\begin{tabular}{lcccc}
\toprule

\multirow{2}{*}{\textbf{Model}} 
& \multicolumn{2}{c}{\textbf{Replica}} 
& \multicolumn{2}{c}{\textbf{Matterport3D}} \\
\cmidrule(lr){2-3}\cmidrule(lr){4-5}
& SR $\uparrow$ & SPL $\uparrow$
& SR $\uparrow$ & SPL $\uparrow$ \\
\midrule
w/o TFM and w/o CPP & 91.4 & 74.4 & 67.7 & 54.3 \\
w/o TFM & 85.4 & 75.9 & 69.0 & 59.4 \\
w/o CPP & 56.3 & 41.4 & 45.8 & 42.8 \\
\textbf{TDGP (Ours)} & \textbf{98.7} & \textbf{89.2} & \textbf{73.7} & \textbf{62.6} \\ 
\bottomrule 

\end{tabular}%
}

\vspace{2pt}
{\footnotesize \textit{Note: ``w/o'' denotes without.}}
\end{table}

\paragraph{Transition from Local Goals to Atomic Actions}
At each time step $t$, after the planner receives the local goal $\mathbf{g}_t$, it first performs a coordinate transformation and node mapping: It transforms $\mathbf{g}_t$ ($3 \times 3$ grid index) from the agent’s egocentric coordinate system to global map coordinates, and then maps it to the nearest node $\mathbf{v}_g$ on the navigation map $\mathcal{G}_t$. Subsequently, the planner uses the Dijkstra algorithm to compute the shortest path $P$ from the current node $\mathbf{v}_c$ to the target node $\mathbf{v}_g$:

\begin{equation}
P(\mathbf{v}_c, \mathbf{v}_g) = \text{Dijkstra}(\mathcal{G}_t, \mathbf{v}_c, \mathbf{v}_g),
\end{equation}
The planner extracts the next node $\mathbf{v}_n$ from the path $P$. By comparing the relative azimuth between $\mathbf{v}_c$ and $\mathbf{v}_n$ with the agent’s current orientation $\mathbf{\theta}_t$, the required atomic action $\mathbf{a}_t$ is computed:

\begin{equation}
\mathbf{a}_t = \text{Derive}(\mathbf{v}_c, \mathbf{v}_n, \mathbf{\theta}_t).
\end{equation}

\paragraph{Collision-driven graph update mechanism}
\begin{itemize}
    \item Path Planning Map Update: When there are passive visual limitations, and the scene is static, navigation can only be achieved through collisions, guided by sound. While visual sensors may be misled by transparent obstacles or abnormal lighting conditions, physical collisions provide reliable and trustworthy signals. Therefore, when the agent executes action $\mathbf{a}_t$ and receives a collision signal $C_t$ from the environment, the planner identifies the node $\mathbf{v}_n'$ that the agent attempted to reach in the previous step. The planner immediately removes the edge $e$ connecting the current node $\mathbf{v}_c$ and $\mathbf{v}_n'$ from the edge set $E_t$, thereby updating the navigation graph $\mathcal{G}$ at the next time step $t+1$:

    \begin{equation} 
    E_{t+1} = E_t \setminus \{e(\mathbf{v}_c, \mathbf{v}_n') | C_t = \text{True}\} 
    \end{equation}

    Here, $\mathbf{v}_n'$ is the next node in the path planned in the previous step, and $C_t \in \{\text{True}, \text{False}\}$ is the collision indicator signal returned by the environment. This design enables the planner to learn from physical collisions and dynamically refine its internal path planning model.
    \item Collision Penalty: Although graph updates ensure robustness, relying solely on physical collisions ($C_{t}$) may limit efficiency. To address this issue, we introduce a collision penalty term $r_{c}$ into the reward function:
    
    \begin{equation}
    R_{t} = r_{g} + r_{s} + r_{c},
    \end{equation}
    Here, $R_{t}$ represents the total reward, $r_{g}$ is the target reward, and $r_{s}$ is the proximity reward: a positive reward is given when the agent is close to the target, and a negative reward when it is far away. This explicitly incentivizes high-order policies to learn proactive avoidance during training, enabling them to anticipate potential dangers and select safer targets.
\end{itemize}

\begin{table*}[htbp]
\centering
\caption{Audio Enhancement Ablation Study without complex scenarios. The + comp notation in the model column indicates that complex scenes are utilized during training but are not employed during testing/evaluation.}
\label{tab:tab2}
\resizebox{\textwidth}{!}{
\begin{tabular}{@{}l ccc ccc ccc ccc@{}}
\toprule
\multirow{2}{*}{\textbf{Model}} & \multicolumn{6}{c}{\textbf{Replica}} & \multicolumn{6}{c}{\textbf{Matterport3D}} \\
\cmidrule(lr){2-7} \cmidrule(lr){8-13}
 & \multicolumn{3}{c}{Multiple Heard} & \multicolumn{3}{c}{Unheard} & \multicolumn{3}{c}{Multiple Heard} & \multicolumn{3}{c}{Unheard} \\
\cmidrule(lr){2-4} \cmidrule(lr){5-7} \cmidrule(lr){8-10} \cmidrule(lr){11-13}
 & SPL$\uparrow$ & SR$\uparrow$ & SNA$\uparrow$ & SPL$\uparrow$ & SR$\uparrow$ & SNA$\uparrow$ & SPL$\uparrow$ & SR$\uparrow$ & SNA$\uparrow$ & SPL$\uparrow$ & SR$\uparrow$ & SNA$\uparrow$ \\
\midrule
AV-Nav + comp~\cite{b2} & 62.6 & 85.6 & 31.3 & 45.4 & 67.0 & 21.6 & 46.3 & 68.3 & 26.3 & 30.9 & 51.2 & 16.1 \\
SAVi + comp~\cite{b23} & 43.1 & 58.6 & 25.0 & 33.6 & 45.9 & 18.1 & 48.9 & 61.4 & 35.2 & 35.9 & 47.5 & 25.4 \\
DB-Nav + comp~\cite{b17} & 55.7 & 77.4 & 39.4 & \textbf{54.0} & 77.8 & 37.4 & \textbf{63.8} & 83.5 & 47.0 & \textbf{49.9} & 70.6 & 35.8 \\
\textbf{Ours + comp} & \textbf{74.5} & 85.2 & 56.0 & 48.4 & 58.6 & 34.3 & 59.7 & 73.3 & 50.0 & 38.7 & 49.7 & 28.9  \\
\bottomrule
\end{tabular}%
}
\end{table*}

\subsection{Audio Enhancement}

To improve the robustness and generalization ability of the policy network in complex acoustic environments~\cite{mattursun2024bss,zhang2024nonlinear,cao2024vnet}, we employ the audio data augmentation strategy proposed by Younes et al.~\cite{b17}. During training, the simulator probabilistically activates one or more augmentations at the start of each episode according to the configuration. A noise source can be generated at a random non-target location and mixed with the target sound, while a second sound source may also be mixed at the target location. Spectral masking further applies temporal and frequency masks to the generated spectrogram. Multiple augmentation settings are evaluated in our experiments.

\begin{table*}[htbp]
\centering
\caption{Audio Enhancement Ablation Study with complex scenarios.}
\label{tab:tab4}
\resizebox{\textwidth}{!}{
\begin{tabular}{@{}l ccc ccc ccc ccc@{}}
\toprule
\multirow{2}{*}{\textbf{Model}} & \multicolumn{6}{c}{\textbf{Replica}} & \multicolumn{6}{c}{\textbf{Matterport3D}} \\
\cmidrule(lr){2-7} \cmidrule(lr){8-13}
 & \multicolumn{3}{c}{Multiple Heard} & \multicolumn{3}{c}{Unheard} & \multicolumn{3}{c}{Multiple Heard} & \multicolumn{3}{c}{Unheard} \\
\cmidrule(lr){2-4} \cmidrule(lr){5-7} \cmidrule(lr){8-10} \cmidrule(lr){11-13}
 & SPL$\uparrow$ & SR$\uparrow$ & SNA$\uparrow$ & SPL$\uparrow$ & SR$\uparrow$ & SNA$\uparrow$ & SPL$\uparrow$ & SR$\uparrow$ & SNA$\uparrow$ & SPL$\uparrow$ & SR$\uparrow$ & SNA$\uparrow$ \\
\midrule
AV-Nav + comp~\cite{b2} & 53.8 & 80.1 & 26.5 & 44.4 & 67.7 & 21.7 & 41.5 & 66.3 & 22.5 & 33.3 & 55.5 & 17.6 \\
SAVi + comp~\cite{b23} & 34.9 & 51.0 & 20.4 & 30.0 & 42.8 & 16.6 & 44.2 & 59.9 & 31.8 & 37.6 & 52.9 & 26.6 \\
DB-Nav + comp~\cite{b17} & 50.5 & 74.5 & 35.8 & \textbf{49.9} & 74.8 & 35.4 & \textbf{60.1} & 82.2 & 44.4 & \textbf{52.6} & 72.5 & 37.9 \\
Ours + comp & \textbf{68.8} & \textbf{80.5} & \textbf{53.8} & 49.0 & 60.2 & 35.7 & 59.3 & 72.9 & 48.8 &  37.1 & 46.0 & 27.2  \\
\bottomrule
\end{tabular}%
}
\end{table*}

\section{Experiment}
\subsection{Experimental Setup}
All experiments were conducted on the SoundSpaces 1.0 platform. SoundSpaces is built on the Habitat simulator \cite{b2} and provides realistic visual rendering and physics-based audio simulation. This makes it an ideal environment for conducting audio-visual navigation research.
During training and evaluation, we utilized two large-scale 3D scene datasets. The Replica dataset \cite{b21} contains 18 real-world indoor scenes captured via high-quality scans, providing detailed geometric and textural information. The Matterport3D (MP3D) \cite{b22} dataset, on the other hand, comprises 85 larger-scale, more diverse indoor environments; this enables us to more effectively evaluate the model’s generalization capabilities across different scenarios. To maintain consistency with previous work, we divided the scenes into three groups: training, validation, and testing, consisting of 9/4/5 scenes for Replica and 73/11/18 scenes for MP3D. We evaluate the performance of our surrogates in both environments—heard and unheard—under the following two distinct settings. The unheard setting in MP3D is the most demanding and is our primary focus.

\begin{figure}[htbp]
\includegraphics[width=1\columnwidth]{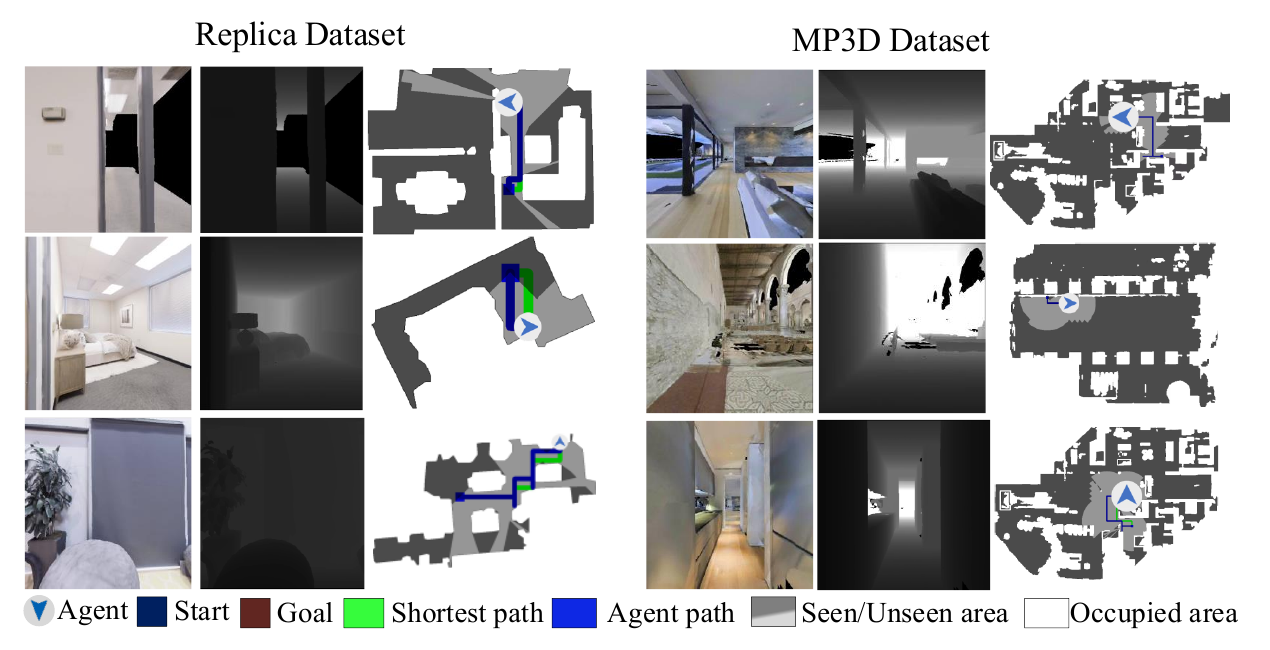} 
\caption{This section presents the navigation environments and model inputs from the Replica and MP3D datasets. For each scene example, the visualizations include the raw $\text{RGB}$ image, the corresponding depth map input, and the final top-down map. The top-down map clearly shows the planning space, the shortest path (green), and the agent’s current position and field of view.}
\label{fig:fig3}
\end{figure}

\begin{figure}[htbp]
\includegraphics[width=1\columnwidth]{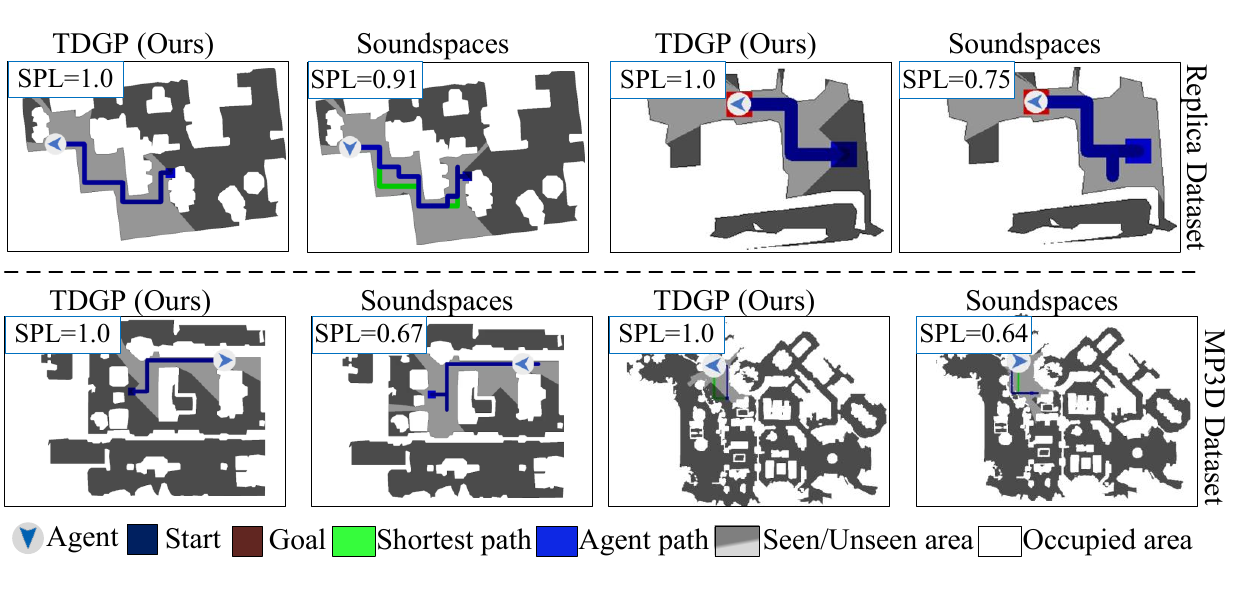} 
\caption{This figure compares our navigation paths and SPL scores with those of the SoundSpaces baseline model on the Replica and MP3D datasets.}
\label{fig:fig4}
\end{figure}

\subsection{Performance of the Transformer-based Token Fusion and Dynamic Graph Planning model}
\paragraph{Model Performance}
Table ~\ref{tab:tab1} shows that, compared to SoundSpaces, our TDGP model achieves improvements of 14.8\%, 7.3\%, and 25.6\% in Success
weighted by Path Length(SPL), Success Rate(SR), and Success weighted by Normalized Agent-Action(SNA), respectively, on the Replica dataset~\cite{b21} in the heard scenario, and improvements of 7.0\%, 9.1\%, and 9.5\% in the unheard scenario. On the MP3D~\cite{b22} dataset, the TDGP model improves SPL, SR, and SNA by 8.3\%, 6.0\%, and 22.9\% in the heard scenario, and by 14.2\%, 8.7\%, and 15.5\% in the unheard scenario.

Although our TDGP model improves the navigation efficiency of audio-visual navigation, it still has several limitations. First, CPP relies on a projected 2D occupancy grid map, which is susceptible to depth-sensor noise and representation inaccuracies in large-scale, complex environments such as MP3D. Consequently, while CPP ensures the agent’s SR, it may sacrifice SPL compared with navigation under ideal, noise-free geometric perception. Second, consistent with mainstream audio-visual navigation benchmarks, our system is optimized for static environments. After detecting a collision, CPP removes the corresponding edge from the navigation graph. While this strategy is effective in static scenes, permanent edge removal may discard paths that become valid again in dynamic real-world environments. To address this, future work will explore time-decay edge recovery, which gradually reconsiders removed edges and reactivates them when they are likely to become traversable again.

\paragraph{Ablation Studies}
Table \ref{tab:tab3} confirms that the multi-level fusion strategy (TFM) and the collision-penalized path planner (CPP) are crucial. The w/o TFM configuration (replacing Transformer fusion with concatenation) hinders cross-modal interaction, thereby reducing multimodal reasoning capabilities. Meanwhile, the w/o CPP configuration (a traditional end-to-end system) leads to unstable navigation because a single network struggles to balance high-level reasoning with fine-grained control. The superior performance of the complete model on the SR and SPL metrics highlights the synergistic effects between deep fusion and hierarchical planning. 
\subsection{A Study on Ablation with Audio Enhancement}
We evaluated the generalization performance of the audio enhancement strategy ($+ \text{comp}$). Tables \ref{tab:tab2} and \ref{tab:tab4} demonstrate the robustness of this strategy in unseen acoustic scenarios. Our TDGP model exhibits a performance gap compared to DB-Nav, which is specifically designed to handle complex and even moving sound source models. DB-Nav relies on specific recurrent neural networks and complex acoustic memory banks to fit audio noise. In contrast, the core of our TDGP model lies in its high-level token fusion mechanism and collision-penalty graph planner. Specifically, a comparison of Tables \ref{tab:tab1} and \ref{tab:tab4} reveals that the SR on the MP3D unseen dataset improved from 42.2\% to 46.0\%, a 3.8\% increase, indicating that our model possesses practical value in noisy scenarios.

\begin{figure}[htbp]
\includegraphics[width=1\columnwidth]{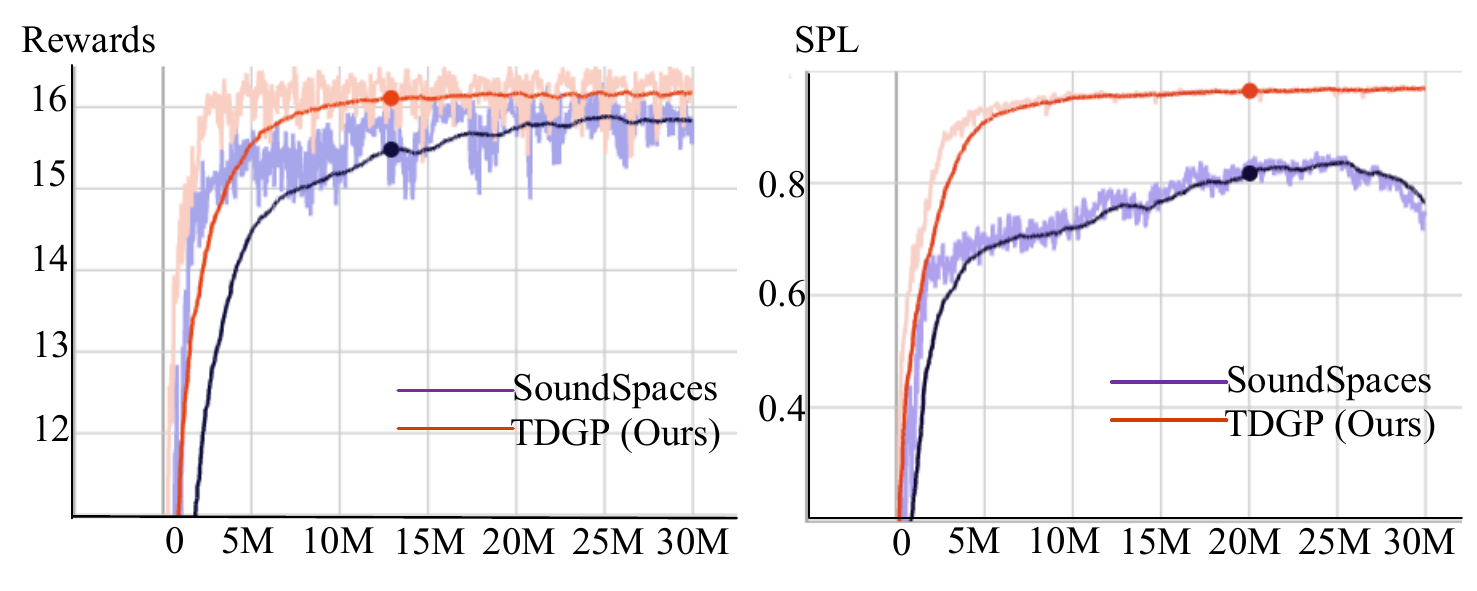} 
\caption{Qualitative Analysis. The navigation metrics vary with the training steps.}
\label{fig:fig5}
\end{figure}

\subsection {Qualitative Analysis}
To gain a more intuitive understanding of how our model works, we conducted a qualitative analysis of several typical navigation trajectories. Figs. \ref{fig:fig3} and \ref{fig:fig4} show navigation trajectories on the top-down map of the TDGP model, effectively guiding the agent to the target. Fig. \ref{fig:fig5} illustrates how the navigation metric changes with the number of training steps, achieving a higher SPL value with fewer training steps, indicating that it converges faster than SoundSpaces.
\section{Conclusion}
To address the deceptive and incomplete nature of passive vision in audio-visual navigation, this paper proposes TDGP, a Transformer-based Token Fusion and Dynamic Graph Planning model. By combining token-level multimodal fusion with a collision-penalty graph planner, TDGP enables robust local decision-making and adaptive path replanning under perceptual uncertainty. Experiments on Replica and Matterport3D demonstrate its effectiveness in path efficiency and generalization, especially in unheard acoustic scenarios. Nevertheless, the current evaluation is conducted in simulation, and real-world deployment still requires handling sensor calibration errors, actuator noise, latency, and dynamic obstacles. Future work will focus on active geometric planning and physical robotic deployment to bridge the gap between simulated perception and real-world navigation.

\bibliographystyle{IEEEtran}
\bibliography{references}

\end{document}